%% file: paper_arxiv.tex
\documentclass[10pt,twocolumn,letterpaper]{article}

\usepackage{cvpr}              %

\usepackage{graphicx}
\usepackage{amsmath}
\usepackage{amssymb}
\usepackage{booktabs}
\usepackage[utf8]{inputenc} %
\usepackage[T1]{fontenc}    %
\usepackage{multirow}
\usepackage{pifont}
\usepackage[accsupp]{axessibility}

\usepackage[pagebackref,breaklinks,colorlinks]{hyperref}

\usepackage[capitalize]{cleveref}

\crefname{section}{Sec.}{Secs.}
\Crefname{section}{Section}{Sections}
\Crefname{table}{Table}{Tables}
\crefname{table}{Tab.}{Tabs.}

\def\cvprPaperID{10385} %
\def\confName{CVPR}
\def\confYear{2022}

\include{shortcuts}

\begin{document}

\title{Trajectory Optimization for Physics-Based Reconstruction of \\ 3d Human Pose from Monocular Video}

\author{Erik Gärtner\textsuperscript{\rm 1,}\textsuperscript{\rm 2}\\
\and
Mykhaylo Andriluka\textsuperscript{\rm 1}\\
\and
Hongyi Xu\textsuperscript{\rm 1 }
\and
Cristian Sminchisescu\textsuperscript{\rm 1}
\and
\textsuperscript{\rm 1}\bf{Google Research}, \textsuperscript{\rm 2}\bf{Lund University}\\
{\tt\small erik.gartner@math.lth.se}\\
{\tt\small \{mykhayloa,hongyixu,sminchisescu\}@google.com}
}

\maketitle

\input{abstract_new}

\input{intro_new}

\input{related_work}

\input{model}

\input{experiments_arxiv}

\input{conclusion}

\clearpage
{\small
\bibliographystyle{ieee_fullname}
\bibliography{biblio}
}

\clearpage
\input{supplementary}

\end{document}

%% file: shortcuts.tex
\newcommand{\cc}{\mathbf{c}}

\newcommand{\xx}{\mathbf{x}}

\newcommand{\bvl}{\mathbf{l}}

\newcommand{\ee}{\mathbf{e}}
\newcommand{\vv}{\mathbf{v}}

\newcommand{\qq}{\mathbf{q}}
\newcommand{\qqd}{\mathbf{\dot q}}

\newcommand{\zz}{\mathbf{z}}

\newcommand{\CC}{\mathbf{C}}

\newcommand{\BB}{\mathbf{B}}
\newcommand{\TT}{\mathbf{T}}

\newcommand{\MM}{\mathbf{M}}

\newcommand{\JJ}{\mathbf{J}}

\newcommand{\bbeta}{\boldsymbol{\beta}}
\newcommand{\btheta}{\boldsymbol{\theta}}

\newcommand{\bpsi}{\boldsymbol{\psi}}

\newcommand{\btau}{\boldsymbol{\tau}}

\newcommand{\myparagraph}[1]{\vspace{0.1em}\noindent\textbf{#1}}

\newcommand{\Eq}[1]{\eqref{#1}}
\newcommand{\Figure}[1]{fig.~\ref{#1}}
\newcommand{\Table}[1]{tab.~\ref{#1}}
\newcommand{\Section}[1]{\S\ref{#1}}

\newcommand{\mpp}{\emph{(mp)}}

%% file: abstract_new.tex
\begin{abstract}
We focus on the task of estimating a physically plausible articulated human motion from monocular video.
Existing approaches that do not consider physics often produce temporally inconsistent output with
motion artifacts, while state-of-the-art physics-based approaches have either been shown to work only in
controlled laboratory conditions or consider simplified body-ground contact limited to feet.
This paper explores how these shortcomings can be addressed by directly incorporating
a fully-featured physics engine into the pose estimation process. Given an uncontrolled,
real-world scene as input, our approach estimates the ground-plane location and the dimensions of the
physical body model. It then recovers the physical motion by performing trajectory optimization.
The advantage of our formulation is that
it readily generalizes to a
variety of scenes that might have diverse ground properties and supports any
form of self-contact and contact between the articulated body and scene geometry.
We show that our approach achieves competitive results with respect to existing physics-based
methods on the Human3.6M benchmark \cite{h36m_pami}, while being directly applicable without 
re-training to more complex dynamic motions from the AIST benchmark \cite{aist-dance-db} and to uncontrolled
internet videos.
\end{abstract}

%% file: intro_new.tex
\section{Introduction}
\label{sec:intro}
\input{fig_teaser}
\input{fig_teaser_part2}

In this paper, we address the challenge of reconstructing physically plausible articulated 3d human
motion from monocular video aiming to complement the recent methods
\cite{kanazawa2018end,xiang2019monocular,zanfir2020neural, pavllo:videopose3d:2019,xiang2019monocular,kocabas20cvpr} that achieve increasingly
more accurate 3d pose estimation results in terms of standard joint accuracy metrics, but still
often produce reconstructions that are visually unnatural.

Our primary mechanism to achieve physical plausibility is to incorporate laws of physics into the
pose estimation process. This naturally allows us to impose a variety of desirable properties on the estimated articulated motion, such as temporal consistency and balance in the presence of
gravity. Perhaps one of the key challenges in using physics for pose estimation is the inherent
complexity of adequately modeling the diverse physical phenomena that arise due to interactions of
people with the scene.
In the recent literature \cite{RempeContactDynamics2020,PhysCapTOG2020,PhysAwareTOG2021,xie2021iccv}
it is common to keep the physics model simple to enable efficient inference. For example, most of the recent approaches opt for using simplified contact models (considering foot contact only), ignore
potential effects due to interaction with objects other than the ground-plane, and do not model more
subtle physical effects such as sliding and rolling friction, or surfaces with varying degrees of
softness. Clearly there are many real-world scenarios where leveraging a more feature-complete physical model is necessary.
We explore physics-based articulated pose estimation
using feature-complete physical simulation as a building block to address this shortcoming. The advantage of such an approach is
that it allows our method to be readily applicable to a variety of motions and scenarios that have not
previously been tackled in the literature (see fig.~\ref{fig:teaser} and
\ref{fig:teaser2}). Specifically, in contrast to \cite{RempeContactDynamics2020, PhysCapTOG2020,
  PhysAwareTOG2021, xie2021iccv} our approach can reconstruct motions with any type of contact
between the body and the ground plane (see fig.~\ref{fig:teaser}). Our approach can also model
interaction with obstacles and supporting surfaces such as furniture and allows for varying the stiffness
and damping of the ground-plane to represent special cases such as trampoline floor (see
fig.~\ref{fig:teaser2}). We rely on the Bullet \cite{coumans2019pybullet} engine, which was
previously used for simulating human motion in \cite{peng19dm}. However, none of our implementation
details are engine-specific, so we envision that the quality of our results might continue to improve
with further development in physical simulation.

The main contribution of this paper is to experimentally evaluate the use of trajectory optimization
for physics-based articulated motion estimation on laboratory and real-world data using a generic
physics engine as a building block. We demonstrate that combining a feature-complete physics
engine and trajectory optimization can reach competitive or better accuracy than
state-of-the-art methods while being applicable to a large variety of scenes and motion types. Furthermore, to the
best of our knowledge, we are the first to apply physics-based reconstruction to complex
real-world motions such as the ones shown in fig.~\ref{fig:teaser} and \ref{fig:teaser2}.
As a second contribution, we generate technical insights such as demonstrating that we can reach
excellent alignment of estimated physical motion with 2d input images by automatically adapting the
3d model to the person in the image, and employing appropriate 2d alignment losses. This is in
contrast to related work \cite{RempeContactDynamics2020, PhysCapTOG2020, PhysAwareTOG2021,
  xie2021iccv} that typically does not report 2d alignment error and qualitatively may not achieve good
2d alignment of the physical model with the image. We also contribute to the understanding of the
use of the residual root force control \cite{yuan2020residual}. Such residual root force has been
hypothesized as essential to bridge the simulation-to-reality gap and compensate for
inaccuracies in the physical model. We experimentally demonstrate that the use of physically unrealistic residual force control might not be necessary, even in cases of complex and dynamic motions.

%% file: fig_teaser.tex
\begin{figure}
  \begin{center}
    \includegraphics[width=\linewidth]{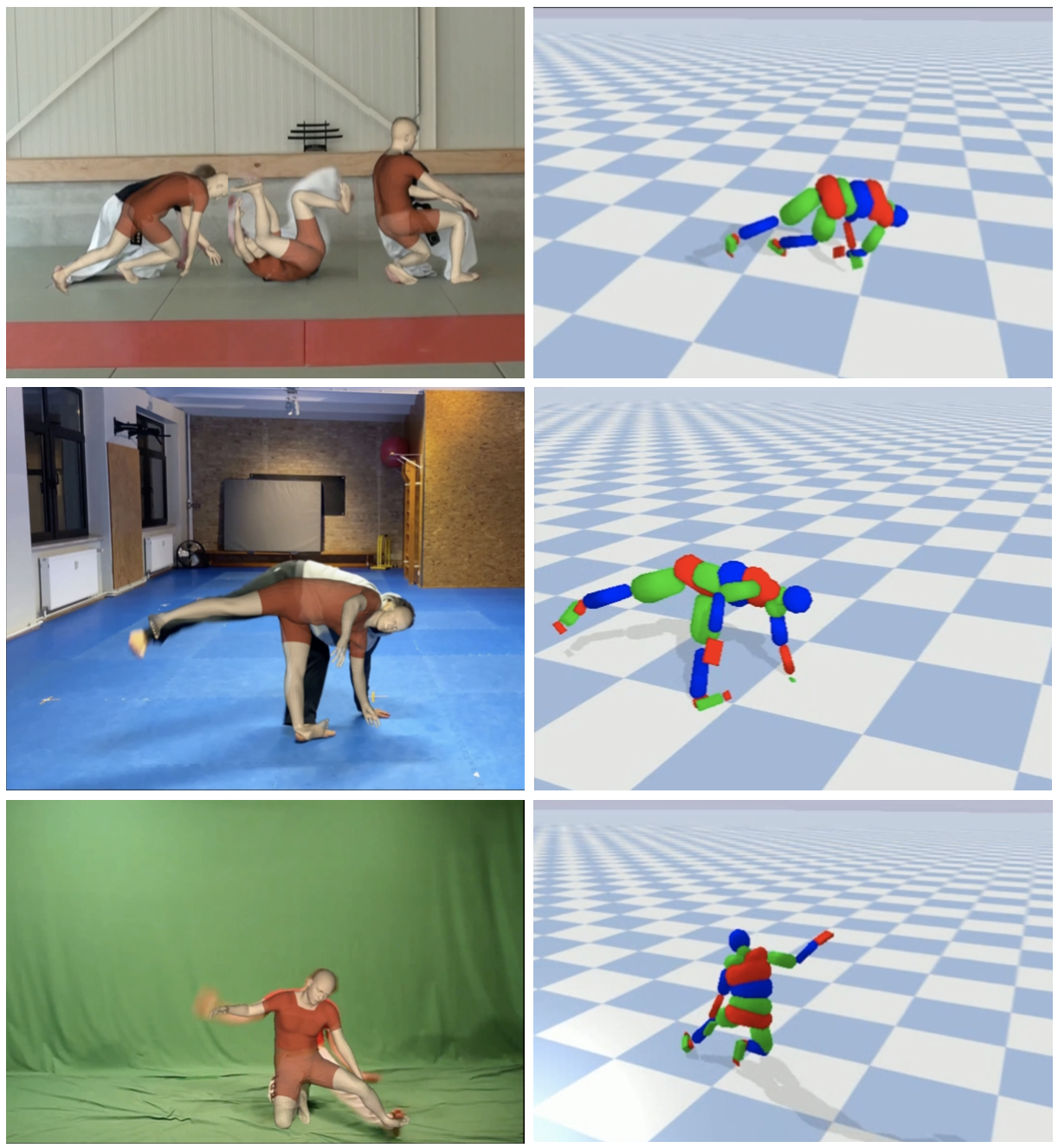}
  \caption{Example results of our approach on internet videos of dynamic motions. Note that our model can reconstruct physically
    plausible articulated 3d motion even in the presence of complex contact with the
    ground: full body contact (top row), feet and hands (middle), and feet and knee contacts (bottom).}
    \label{fig:teaser}
  \end{center}
\vspace{-7mm}
\end{figure}

%% file: fig_teaser_part2.tex
\begin{figure}
  \begin{center}
    \includegraphics[width=0.93\linewidth]{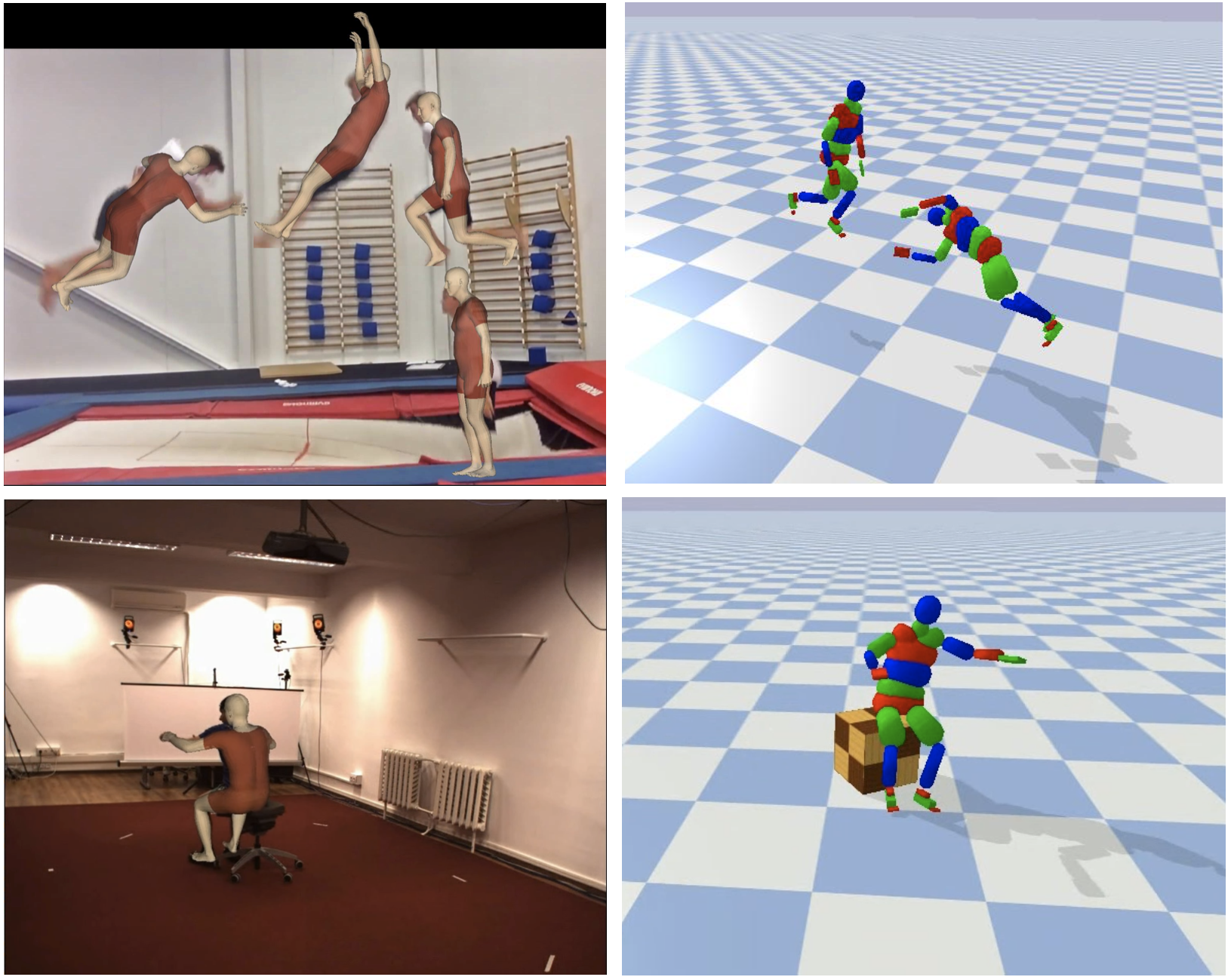}
    \vspace{-2mm}
    \caption{Examples results of our approach for scene with soft ground (top) and interaction with
      a chair (bottom).}
    \label{fig:teaser2}
  \end{center}
\vspace{-8mm}
\end{figure}

%% file: related_work.tex
\section{Related work}
\label{sec:relatedwork}
\input{fig_overview}
\input{tab_compare_physics}

In the following, we first discuss recent literature on 3d human pose estimation that does not
incorporate physical reasoning. We then review the related work on physics-based human modeling and
compare our approach to other physics-based 3d pose estimation approaches.

\myparagraph{3d pose estimation without physics.}
State-of-the-art methods are highly effective in estimating 2d and 3d people poses in images \cite{cao2017openpose-cvpr, zanfir2018deep, kanazawa2018end}, and recent work has been able to extend this progress to 3d pose estimation in
video \cite{pavllo:videopose3d:2019, xiang2019monocular, kocabas20cvpr}.
The key elements driving the
performance of these methods is the ability to estimate data-driven priors on articulated 3d poses
\cite{kocabas20cvpr,zanfir2020weakly} and learn sophisticated CNN-based representations from large
corpora of annotated training images \cite{h36m_pami,AMASS:ICCV:2019, vonMarcard2018,joo_total_motion_cap}.
As such, these methods perform very well on common poses but are still challenged by rare
poses. Occlusions, difficult imaging conditions, and dynamic motions (e.g. athletics) remain a challenge as
these are highly diverse and hard to represent in the training set. As pointed out in 
\cite{RempeContactDynamics2020}, even for common poses state-of-the-art methods still often
generate reconstructions prone to artifacts such as floating, footskating, and non-physical leaning. We aim to complement the statistical models used in the state-of-the-art approaches by incorporating laws of physics into the inference process and thus adding a component that is universally
applicable to any human motion regardless of the statistics of the training or test set.

In parallel with recent progress in pose estimation, we now have accurate statistical shape and pose models \cite{SMPL:2015, anguelov2005scape,xu2020ghum}. These body models are typically estimated from thousands of scans of people and can generate shape deformations for a given pose. In this paper, we take advantage of these improvements and use a statistical body shape model \cite{xu2020ghum} to define the dimensions of our physical model and derive the mass from the volume of the body parts.

\myparagraph{Physics-based human motion modeling.} 
Human motion modeling has been a subject of active research in computer graphics
\cite{alborno18cgf, lee19tog}, robotics \cite{dasilva2008} and reinforcement learning \cite{heess2017richlocomotion, peng19dm, ScaDiver} literature. With a few exceptions, most of the models in these domains have been constructed and evaluated using the motion capture data \cite{alborno18cgf}. Some work such as \cite{2018-TOG-SFV} use images as input, aiming to train motion controllers for a simulated character capable of performing the observed motion under various perturbations. That work focuses on training motion controllers for a fixed character, whereas our focus is on estimating the motion of the subject observed in the image.  
Furthermore, the character's size, shape, and mass are independent of the observed subject.
\cite{lee19tog} propose a realistic human model that directly represents muscle activations and a method to learn control policies for it. \cite{won2019tog} generate motions for a variety of
character sizes and learn control policies that adapt to each size. \cite{lee19tog,won2019tog} and
similar results in the graphics literature do not demonstrate this for characters observed in real
images and do not deal with challenges of jointly estimating physical motion and coping with
ambiguity in image measurements or the 2d to 3d lifting process~\cite{sminchisescu_cvpr03}.

\myparagraph{Physics-based 3d pose estimation.} 
Physics-based human pose estimation has a long tradition in computer vision
\cite{brubaker09iccv,metaxas97,vondrak08cvpr}. Early works such as \cite{vondrak08cvpr} already
incorporated physical simulation as prior for 3d pose tracking but only considered simple motions
such as walking and mostly evaluated in the multi-view setting in the controlled laboratory
conditions. We list some of the properties of the recent
works in tab.~\ref{tab:comparephys}.
\cite{li2019cvpr} demonstrate joint physics-based estimation of human motion and interaction with various
tool-like objects.  \cite{RempeContactDynamics2020} proposes a formulation that simplifies
physics-based reasoning to feet and torso only, and infers positions of other body parts through
inverse kinematics, whereas
\cite{li2019cvpr} jointly model all body parts and also include forces due to interaction with an
object. 
\cite{PhysCapTOG2020, PhysAwareTOG2021} use a specialized physics-based formulation that solves for
ground-reaction forces given pre-detected foot contacts and kinematic estimates. In contrast, we do
not assume that contacts can be detected a-priori, and in our approach, we estimate these as part of
the physical inference. Hence we are not limited to predefined types of contact
as~\cite{li2019cvpr,RempeContactDynamics2020,PhysCapTOG2020,PhysAwareTOG2021} or their accurate
a-priori estimates. We show that we quantitatively improve over~\cite{RempeContactDynamics2020,PhysCapTOG2020}, and qualitatively show how we can address more difficult in-the-wild internet
videos of activities such as somersaults and sports, which would be difficult to reconstruct using previous methods.
Our work is conceptually similar to SimPoE \cite{yuan2021simpoe} in that both works use physics
simulation. In contrast to SimPoE, we introduce a complete pipeline that is applicable to real-world
videos, whereas SimPoE has been tested only in laboratory conditions and requires a
calibrated camera. Furthermore, since SimPoE relies on reinforcement learning to train dataset-specific neural network models to control the simulated body, it is not clear how well SimPoE would generalize to variable motions present in
real-world videos. One clear advantage of the SimPoE approach is its fast
execution at test time, which comes at the cost of lengthy pre-training. 
Our approach is related to the approach of \cite{xie2021iccv} which also estimates 3d
human motion by minimizing an objective function that incorporates physics constraints. 
Perhaps the most
significant differences to \cite{xie2021iccv} are that (1) we use the full-featured physics model
whereas they consider simplified physical model, (2) their model considers physics-based loss, but
the output is not required to correspond to actual physical motion, and (3) they do
not discuss performance of the approach on real-world data. The advantage of \cite{xie2021iccv} is
that they define a differentiable model that can be readily optimized with gradient descent. Finally, the concurrent work \cite{gartner2022diffphy} tackles physics-based human pose reconstruction by minimizing a loss using a differentiable physics simulator given estimated kinematics.

%% file: fig_overview.tex
\begin{figure*}[!t]
    \begin{center}
    \includegraphics[width=1\linewidth]{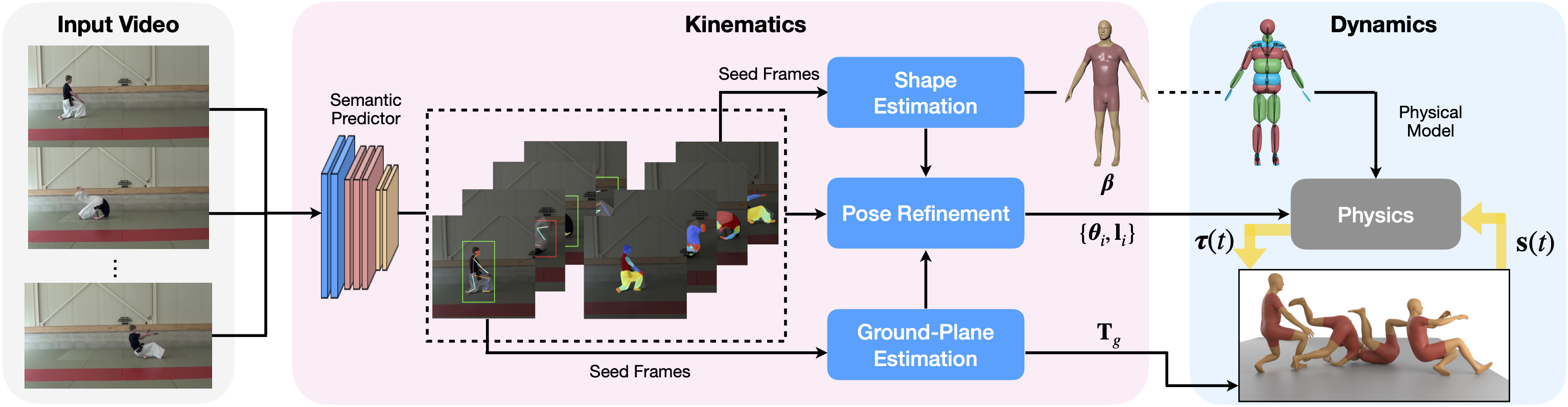}
    \vspace{-4mm}
    \caption{\textbf{Overview.} Given a monocular video of a human motion, we estimate the parameters of a physical human model and motor control trajectories $\btau(t)$ such that the physically simulated human motion aligns with the video. We first use an inference network that predicts 2d landmarks $\mathbf{l}_i$ and body semantic segmentation masks from the video frames. From $n$ seed frames we estimate a time-consistent human shape $\bbeta$ and the ground-plane location $\TT_g$. These are then kept fixed during a per-frame pose refinement step which provides the 3d kinematic initialization $\{\btheta_i\}$ to the physics optimization. The dynamics stage creates a physical model that mirrors the statistical shape model with appropriate shape and mass. Our dynamics optimization improves 3d motion estimation taking into account 3d kinematics, 2d landmarks and physical constraints. We refer to \S\ref{sec:approach} for details.}
    \label{fig:overview}
    \end{center}
    \vspace{-4mm}
\end{figure*}

%% file: tab_compare_physics.tex
\begin{table*}[t]
\begin{small}
\begin{center}
\begin{tabular}{l c c c c c c }
\hline
 & Contact model & Real-time & Physics implementation & Residual force & Body model & Real-world videos \\
\hline
Li \etal \cite{li2019cvpr} & body joints & no & custom & no & fixed & yes \\
Rempe \etal \cite{RempeContactDynamics2020} & feet &  no & custom  & no & fixed & yes \\
PhysCap \cite{PhysCapTOG2020} & feet &  yes & custom  & yes & fixed & yes \\
Shimada \etal \cite{PhysAwareTOG2021} & feet & yes & custom & yes & fixed & yes \\
SimPoE \cite{yuan2021simpoe} & full body  & yes  &  MuJoCo \cite{mujoco} & yes & adapt. & no  \\
Xie \etal \cite{xie2021iccv} & feet  & no & custom  & no & adapt. & no \\
DiffPhy \cite{gartner2022diffphy} & full body  & no & TDS~\cite{heiden2021neuralsim}  & no & adapt. & yes \\
\hline
Ours & full body  & no & Bullet \cite{coumans2019pybullet} &  no & adapt. & yes \\ 
\hline
\end{tabular}
\end{center}
\vspace{-4mm}
\caption{Comparison of recent physics-based articulated pose estimation approaches. ``Contact
  model'' indicates what contact points between body and ground are considered, ``Residual force''
  indicates if the physical model allows application of additional external force to move the person
  (see \cite{yuan2020residual}), ``Body model'' specifies if approach adapts the physical model to
  person in the video, and ``Real-world videos'' specifies if approach has also been evaluated on
  real-world videos or only on videos captured in laboratory conditions.}
\label{tab:comparephys}
\end{small}
\vspace{-5mm}
\end{table*}

%% file: model.tex
\section{Our approach}
\label{sec:approach}

We present an overview of our approach in \Figure{fig:overview}. Given monocular video as input, we first
reconstruct the initial kinematic 3d pose trajectory using a kinematic approach of \cite{zanfir2020neural} and use it to
estimate body shape and the position of the ground plane relative to the camera. Subsequently,
we instantiate a physical person model with body dimensions and weight that match the estimated body
shape. Next, we formulate an objective function that measures the similarity between the motion of the
physical model and image measurements and includes regularization terms that encourage
plausible human poses and penalize jittery motions. Finally, we reconstruct the physical motion by
minimizing this objective function with respect to the joint torque trajectories.
To realize the physical motion, we rely on the implementation of rigid
body dynamics available in Bullet~\cite{coumans2019pybullet}.

\subsection{Body model and control}
\label{subsec:bodymodel}
We model the human body as 
rigid geometric primitives connected by joints. Our model consists of 26 capsules and has 16 3d body joints for a total of 48 degrees of freedom. We rely on a statistical model of human shape
\cite{xu2020ghum} to instantiate our model for a variety of human body types. To that end, given the
3d mesh representing the body shape, we estimate dimensions of the geometric primitives to
approximate the mesh following the approach of \cite{alborno18cgf}. We then compute the mass and inertia of
each primitive based on its volume and estimate the mass based on an anatomical weight distribution
~\cite{plagenhoef1983anatomical} from the statistical human shape dataset CAESAR~\cite{pishchulin17pr}.

We do not model body muscle explicitly and instead actuate the model by directly applying the
torque at the body joints. We denote the vector of
torques applied at time $t$ as $\btau_t$, the angular position, and velocity of each
joint at time $t$ as $\qq_t$ and $\qqd_t$, and the set of 3d Cartesian coordinates of each joint at time $t$
as $\xx_t$.
Similarly to \cite{2018-TOG-deepMimic}, we control the motion of the physical model by introducing a sequence of control targets $\hat\qq_{1:T} = \{\hat\qq_1, \hat\qq_2, \ldots, \hat\qq_t\}$ which are used to derive the torques via a control loop.
The body motion in our model is then specified by the
initial body state $\mathbf{s}_0= (\qq_{0}, \qqd_{0})$, the world geometry $\mathbf{G}$ specifying the position and
orientation of the ground plane, the control trajectory for each joint $\hat\qq_{1:T}$ and the
corresponding control rule. We assume the initial acceleration to be $\mathbf{0}$.
To implement the control loop we rely on the articulated islands algorithm\footnote{``POSITION\_CONTROL'' mode in Bullet.} (AIA) \cite{stepien2013thesis} 
that incorporates motor control targets as constraints in the linear complementarity problem (LCP)
(\cf (6.3) a, b in \cite{stepien2013thesis}) alongside contact constraints. 
AIA enables stable simulation already at $100$ Hz compared to $1000$-$2000$ Hz for PD control
used in \cite{2018-TOG-deepMimic,alborno18cgf,gartner2022diffphy}.

\subsection{Physics-based articulated motion estimation}
\label{subsec:dynamics}
Our approach to the task of physical motion estimation is generally similar to other trajectory and
spacetime optimization approaches in the literature~\cite{witkin88siggraph,alborno13vcg,alborno18cgf}. We perform optimization over a sequence of
overlapping temporal windows, initializing the start of each subsequent window with the preceding
state in the previous window. To reduce the dimensionality of the search space, we use cubic B-spline
interpolation to represent the control target $\hat\qq_{1:T}$ and perform optimization over the
spline coefficients~\cite{cohen92siggraph}. 
Given the objective function $L$ introduced in \Section{subsec:objfunc} we
aim to find the optimal motion by minimizing $L$ with respect to the spline coefficients of the
control trajectory $\hat{\qq}_{1:T}$. We initialize the control trajectory with the kinematic
estimates of the body joints (see \Section{subsec:kinematic}). The initial state %
is initialized from the corresponding
kinematic estimate. We use the finite difference computed on the kinematic motion to estimate the initial velocity.
As in \cite{alborno13vcg,alborno18cgf} we minimize the objective function with the evolutionary optimization approach CMA-ES~\cite{Hansen2006} since our simulation
environment does not support differentiation with respect to the dynamics variables. We generally observe convergence with CMA-ES after $2000$ iterations per window with $100$ samples per iteration. The inference takes $20-30$ minutes when evaluating $100$ samples in parallel.

\subsection{Objective functions}
\label{subsec:objfunc}
We use a composite objective function given by a weighted combination of several components.

\myparagraph{3d pose.}
To encourage reconstructed physical motion to be close to the estimated kinematic 3d poses
$\qq^{\mbox{\scriptsize{k}}}_{1:T}$ we use the following objective functions
\begin{eqnarray}
    L_{COM}(\hat\qq_{1:T}) &= & \sum_t(\|\cc_t - \cc^{\mbox{\scriptsize{k}}}_t\|^2_2 + \|\dot{\cc_t} -   \dot{\cc}^{\mbox{\scriptsize{k}}}_t\|^2_2) \label{eq:com_loss}\\
    L_{pose} &= &\sum_{t}\sum_{j\in\JJ}\mbox{arccos}(|\langle\qq_{tj}, \qq_{tj}^{\mbox{\scriptsize{k}}}\rangle|)  \label{eq:quat_loss}
\end{eqnarray}
\noindent where $\cc_t$ and $\cc^{\mbox{\scriptsize{k}}}_t$ denote the position of the center of
mass at time $t$ in the reconstructed motion and kinematic estimate. $L_{pose}$ measures the
angle between observed joint angles and their kinematic estimates and the summation 
~\eqref{eq:quat_loss} is over the set $J$ of all body joints including the base joint which defines
the global orientation of the body. 

\myparagraph{2d re-projection.}
To encourage alignment of 3d motion with image observations, we use a set of $N=28$ 
landmark points
that include the main body joints, eyes, ears, nose, fingers, and endpoints of the feet.   
Let $\bvl_{t}$ denote the positions of 3d landmarks on the human body at time $t$, $\CC$ be the
camera projection matrix that maps world points into the image via perspective projection, $\bvl_{t}^d$ be the vector of
landmark detections by the CNN-detector, and $\mathbf{s}_{t}$ the corresponding detection score vector. The 2d
landmark re-projection loss is then defined as
\begin{gather}
\label{eq:keypointloss}
L_{2d} = \sum_{t}\sum_{n}\mathbf{s}_{tn}\|\CC\bvl_{tn} - \bvl_{tn}^{d}\|_2 .
\end{gather}
See \Section{subsec:kinematic} for details on estimating the 2d landmarks.

\myparagraph{Regularization.}
We include several regularizers into our objective function. Firstly, we use the normalizing
flow prior on human poses introduced in \cite{zanfir2020weakly} which penalize unnatural poses. The loss is given by
\begin{gather}
\label{eq:nfprior}
  L_{nf}=\sum_t\|\zz(\qq_t)\|_2 ,
\end{gather}
\noindent where $\zz(\qq_t)$ is the latent code corresponding to the body pose $\qq_t$. 
To discourage jittery motions we a add total variation loss on the acceleration of joints
\begin{gather}
\label{eq:tvloss}
L_{TV} =\frac{1}{J}\sum_t \sum_j\|\ddot{\xx}_{tj} - \ddot{\xx}_{t-1,j}\|_1
\end{gather}
Finally, we include a $L_{lim}$ term that adds exponential penalty on deviations from
anthropomorphic joint limits.
The overall objective $L$ used in physics-based motion estimation is
given by the weighted sum of (\ref{eq:com_loss}- \ref{eq:tvloss}) and of the term $L_{lim}$. See the supplemental material for details.

\subsection{Kinematic 3d pose and shape estimation}
\label{subsec:kinematic}
\input{tab_eval_kinematics_h36m}

In this section, we describe our approach to extracting 2d and 3d evidence from the input video sequence. %

\myparagraph{Body shape.}
Given the input sequence, we proceed first to extract initial per-frame kinematic estimates of the 3d pose and shape using HUND \cite{zanfir2020neural}. As part of its optimization pipeline HUND also recovers the camera intrinsics $\mathbf{c}$ and estimates the positions of 2d landmarks, which we use in the 2d re-projection objective in \Eq{eq:keypointloss}. 
HUND is designed to work on single images, so our initial shape and pose estimates are not temporally consistent. Therefore, to improve the quality of kinematic 3d pose initialization, we extend HUND to pose estimation in
video. We evaluate the additional steps introduced in this section in the experiments shown in
tab.~\ref{tab:videoterms} using a validation set of $20$ sequences from Human3.6M dataset.
In our adaptation, we do not re-train the HUND neural network predictor
and instead, directly minimize the HUND loss functions with BFGS.
As a first step, we re-estimate the shape jointly over multiple
video frames. To keep optimization tractable, we first jointly estimate shape and pose over a subset
of $n=5$ seed frames and then re-estimate the pose in all video frames keeping the updated shape
fixed. The seed frames are selected by the highest average 2d keypoint confidence score.
We refer to the HUND approach with re-estimated shape as \emph{HUND+S} and to our approach where we
subsequently also re-estimate the pose as \emph{HUND+SO}.  In tab.~\ref{tab:videoterms} we show results
for both variants. Note that \emph{HUND+SO} improves considerably compared to the original HUND
results.

\myparagraph{Ground plane.} We define the location of the ground plane by the homogeneous transformation $\TT_g$ that maps from the HUND
coordinates to the canonical coordinate system in which the ground plane is passing through the origin, and its normal is given by the ``y'' axis. Let $\MM^t$ be a subset of points on the body mesh at frame
$t$. The signed distance from the mesh points to the ground plane is given by $D(\MM^t) = \TT_g\MM^t\ee_y$,
where $\ee_y = [0, 1, 0, 0]^T$ is the unit vector of the ``y'' axis in homogeneous coordinates.
To estimate the transformation $\TT_g$ we introduce an objective function
\begin{gather}
L_{gp}(\TT_g, \MM) = \sum_t\|\min(\delta, L_k(D(\MM^t)))\|_2,
\label{eq:gpc}
\end{gather}
where $L_k(D^t)$ corresponds to the smallest $k=20$ signed distances in $D^t$. This objective favors
$\TT_g$ that places body mesh in contact with the ground without making preference for a specific
contact points. This objective is also robust to cases when person is in the air by clipping the
distance at $\delta$, which we set to $0.2$m in the experiments in this paper.
We recover $\TT_g$ by minimizing
\begin{equation}
\begin{split}
  L_{gp}(\TT_g)  = & L_{gp}(\TT_g, \MM_{l}) + L_{gp}(\TT_g, \MM_{r}) \\
  &+ 2L_{gp}(\TT_g, \MM_{b}), 
\end{split}
\end{equation}
where $\MM_{l}$, $\MM_{r}$ and $\MM_{b}$ are the meshes of the left foot, right foot and whole body
respectively. This biases the ground plane to have contact with the feet, but is
still robust to cases when person is jumping or touching the ground with other body parts (e.g. as in the case of a somersault).

\myparagraph{3d pose.}
In the final step, we re-estimate the poses in all frames using the estimated shape and ground plane while adding the temporal consistency objective
\begin{equation}
\label{eq:smoothness}
    L_{temp} = \sum_{t}\|\MM^{t}-\MM^{t-1}\|_2 + 
    \|\btheta_t-\btheta_{t-1}\|_2,
\end{equation}
where $\MM^{t}$ is a body mesh and $\btheta_t$ is a HUND body pose vector in frame $t$.
To enforce ground plane constraints we use ~\eqref{eq:gpc}, but now keep $\TT_g$ fixed and
optimize with respect to body pose. In the experiments in tab.~\ref{tab:videoterms} we refer to the variant of our
approach that uses temporal constraints in ~\eqref{eq:smoothness} as \emph{HUND+SO+T} and to the full
kinematic optimization that uses both temporal and ground plane constraints as \emph{HUND+SO+GT}. Tab.~\ref{tab:videoterms} demonstrates that both temporal and ground-truth constraints considerably
improve the accuracy of kinematic 3d pose estimation. Even so, the results of our best variant 
\emph{HUND+SO+GT} still contain artifacts such as motion jitter and footskating, which are
substantially reduced by the dynamical model (see tab.~\ref{tab:dynablation}).

%% file: tab_eval_kinematics_h36m.tex
\begin{table}[bt]
\begin{small}
\vspace{-1mm}
\begin{center}
\scalebox{1}{
\begin{tabular}{l|c|c|c}
\textbf{Model} & \textbf{MPJPE-G} &\textbf{MPJPE} & \textbf{MPJPE-PA} \\
\hline

HUND~\cite{zanfir2020neural} & 239 & 116 & 72  \\
\phantom{x} + S & 233 & 110 & 71  \\
\phantom{x} + SO & 178 & 85 & 62  \\
\phantom{x} + SO + G & 148 & 84 & 63 \\
\phantom{x} + SO + T & 186 & 85 & 61  \\
\phantom{x} + SO + GT & 135 & 80 & 58 \\
\end{tabular}
}
\end{center}
\vspace{-4mm}
\caption{Ablation of kinematics improvements on HUND on a validation subset of Human3.6M. \emph{+S} indicates time-consistent body shape, \emph{+O} indicates additional non-linear optimization, \emph{+G} using ground-plane constraints, and \emph{+T} temporal smoothness constraints.}

\label{tab:videoterms}
\end{small}
\vspace{-4mm}
\end{table}

%% file: experiments_arxiv.tex
\section{Experimental results}
\label{sec:experiments}
\paragraph{Datasets.}
We evaluate our method on three human motion datasets: Human3.6M~\cite{h36m_pami},
HumanEva-I~\cite{Sigal:IJCV:10b} and AIST~\cite{aist-dance-db}.  In addition, we qualitatively
evaluate on our own ``in-the-wild'' internet videos.  To compare different variants of our approach
in tab.~\ref{tab:videoterms} and tab.~\ref{tab:dynablation} we use a validation set composed of $20$
short 100-frame sequences from the Human3.6M dataset. We use the same subset of full-length
sequences as proposed in \cite{xie2021iccv} for the main evaluation in
tab.~\ref{t:quantitative-all}.
We use a preprocessed version of the AIST dataset \cite{aist-dance-db} from \cite{li2021aistplusplus} 
which contains pseudo 3d body pose ground-truth obtained through multi-view reconstruction.  For our
experiments, we select a subset of fifteen videos featuring diverse dances of single subjects.
For the evaluation on HumanEva-I, we follow the protocol defined in
\cite{RempeContactDynamics2020} and evaluate on the walking motions from the validation split of the
dataset using images from the first camera.
We assume known camera extrinsic parameters in the Human3.6M experiments and estimate them for other
datasets. In order to speed up the computation of the long sequences of Human3.6M in
\Table{t:quantitative-all} we compute all temporal windows in parallel and join them together in post-processing.

We report results using mean global per-joint position error (mm) overall joints (MPJPE-G), as well
as translation aligned (MPJPE) and Procrustes aligned (MPJPE-PA) error metrics.
Note that to score on the MPJPE-G metric an approach should be able to both 
estimate the articulated pose and correctly track the global position of the person in world coordinates.
In addition to standard evaluation metrics, we implement the foot skate and floating metrics similar to those introduced in
\cite{RempeContactDynamics2020} but detect contacts using a threshold rather than through contact annotation. Finally, we report image alignment (MPJPE-2d) and 3d joint velocity error in m/s. See supplementary for further details.

\input{fig_h36m_qualitative}
\input{tab_ablation}

\input{tab_eval_humaneva_aist_h36m}

\input{fig_pim_qualitative_arxiv}

\myparagraph{Analysis of model components.}
In tab.~\ref{tab:dynablation} we present ablation results of our approach. Our full dynamical model uses
kinematic inputs obtained with \emph{HUND+SO+GT} introduced in \S\ref{subsec:kinematic} and is
denoted as \emph{HUND+SO+GT + Dynamics}.  Our dynamical model performs comparably or slightly better
compared to \emph{HUND+SO+GT} on joint localization metrics (e.g. MPJPE-G improves slightly from $135$
to $132$ mm) but greatly reduces motion artifacts. The percentage of frames with
footskate is reduced from $64$ to $8$ and error in velocity from $0.58$ to $0.27$ m/s. We also evaluate a dynamic model based on a simpler kinematic variant
\emph{HUND+SO} that does not incorporate ground-plane and temporal constraints when re-estimating poses
from video. For \emph{HUND+SO}, the inference with dynamics similarly improves perceptual metrics
considerably. Note that \emph{HUND+SO} produces output that suffers from both footskating (25\% of
frames) and floating (40\% of frames). Adding ground-plane constraints in (\cf \Eq{eq:gpc}) removes
floating artifacts in \emph{HUND+SO+GT}, but the output still suffers from footskating (64\% of the
frames). Dynamical inference helps to substantially reduce both types of artifacts both for \emph{HUND+SO} and \emph{HUND+SO+GT}. In fig.~\ref{fig:h36m_example} we show
example output of \emph{HUND+SO+GT + Dynamics}  and compare it to \emph{HUND+SO+GT} which it
uses for initialization. Note that for \emph{HUND+SO+GT} the person in the output appears to move
forward by floating in the air, whereas our dynamics approach infers plausible 3d poses
consistent with the subject's global motion.
In the bottom part of tab.~\ref{tab:dynablation} we report results for our full model \emph{HUND+SO+GT + Dynamics} while ablating components of the objective function
(\cf \Section{subsec:objfunc}).  We observe that all components of the objective function
contribute to the overall accuracy. The most important components are the 2d re-projection
(\cf \Eq{eq:keypointloss}) and difference in COM position (\cf \Eq{eq:com_loss}). Without these, the MPJPE-G increases from $132$ to $154$ and $151$ mm, respectively.  Excluding the 3d joints
component leads to only a small loss of accuracy from $132$ to $134$ mm.

\myparagraph{Comparison to state-of-the-art.}
In tab.~\ref{t:quantitative-all} we present the results of our full model on the Human3.6M, \mbox{HumanEva-I}, and AIST datasets. 
We compare to VIBE~\cite{kocabas20cvpr} using the publicly available implementation by the authors and use the
evaluation results of other approaches as reported in
the original publications. Since VIBE generates only root-relative pose estimates, we use a similar
technique as proposed in PhysCap \cite{PhysCapTOG2020} and estimate the global position and
orientation by minimizing the 2d joint reprojection error.
On the Human3.6M benchmark, our approach improves over VIBE and our own \emph{HUND+SO+GT} in terms of joint
accuracy and perceptual metrics. Compared to VIBE, the MPJPE-G improves from $208$ to $143$
mm, MPJPE-2d improves from $16$ to $13$ px, and the percentage of footskating frames are reduced from
$27\%$ to $4\%$. Interestingly our approach achieves the best MPJPE-PA overall physics-based
approaches except the pretrained SimPoE, but reaches somewhat higher MPJPE compared to \cite{PhysAwareTOG2021} and fairly recent
work of \cite{xie2021iccv} ($82$ mm vs $68$ mm for \cite{xie2021iccv} and $77$ mm for
\cite{PhysAwareTOG2021}). Note that \cite{xie2021iccv} start with a stronger kinematic baseline ($74$ mm
MPJPE) and that the performance of other approaches might improve as well given
such better kinematic initialization.
Furthermore, our dynamics approach improves over the results of
\cite{RempeContactDynamics2020} on \mbox{HumanEva-I} and achieves significantly better MPJPE-G compared to
\emph{HUND+SO+GT}.  On the AIST dataset, dynamics similarly improves in terms of MPJPE-G, footskating,
and velocity compared to our kinematic initialization.

\myparagraph{Results on real-world internet video.}  
We show example results of our approach on the AIST dataset~\cite{aist-dance-db} in fig.~\ref{fig:aistqualitative} 
and on the real-world internet videos in fig.~\ref{fig:teaser}, \ref{fig:teaser2} and \ref{fig:pimqualitative}. 
To obtain the results with a soft floor shown in fig.~\ref{fig:teaser2}
we manually modify the stiffness and damping floor parameters to mimic the trampoline behavior. The
sequence with the chair from the Human3.6M dataset shown in \Figure{fig:teaser2} (bottom) is generated
by manually adding a chair to the scene since our approach does not perform reasoning about
scene objects.

In \Figure{fig:aistqualitative} we qualitatively compare the output of our full system with physics to our best kinematic approach \emph{HUND+SO+GT}. We strongly encourage the
reader to watch the video in supplemental material\footnote{See \href{https://tiny.cc/traj-opt}{tiny.cc/traj-opt}.} to appreciate the differences between the two
approaches and to see the qualitative comparison to VIBE~\cite{kocabas20cvpr}. We observe that our physics approach is often able to correct out-of-balance
poses produced by \emph{HUND+SO+GT} (\eg second frame in fig.~\ref{fig:aistqualitative}) and substantially
improves temporal coherence of the reconstruction. Note that typically both \emph{HUND+SO+GT} and our
physics-based approach produce outputs that match 2d observations, but the physics-based
approach estimates 3d pose more accurately. For example, in the first sequence in fig.~\ref{fig:pimqualitative} the physics-based model infers the pose that enables the person to
jump in subsequent frames, whereas \emph{HUND+SO+GT} places the left leg at an angle that would
make the jump impossible. Note that the output of the physics-based approach can deviate
significantly from the kinematic initialization (fig.~\ref{fig:pimdress} and second example in fig.~\ref{fig:pimqualitative}.
This is particularly prominent in the fig.~\ref{fig:pimdress} where we show example result on a
difficult sequence where 2d keypoint estimation fails to localize the legs in several frames due to
occlusion by the clothing. Note that in this example our full model with dynamics is able to
generate reasonable sequence of 3d poses despite multiple failures in the kinematic initialization.

\myparagraph{Failure cases of our approach.} We show a few characteristic examples of the failure
cases of our approach in fig.~\ref{fig:pimfail}. Note that our physics-based reconstruction depends
on the kinematic 3d pose estimation for initialization and also uses it in one of the components of
the loss (\cf~eq.~\ref{eq:quat_loss}). Therefore our physics-based approach is likely to fail when
kinematic reconstruction is grossly incorrect (see fig.~\ref{fig:pimfail}~(b)) or
when it fails to estimate position of the limb important to maintain the overall pose (see
fig.~\ref{fig:pimfail}~(a)). 
Our physics-based model might also fail when the estimate of
the ground-plane with respect to the camera is inaccurate.
Note how in fig.~\ref{fig:pimfail} (c) the kinematic estimate positions the standing person at an
angle to the true ground-plane normal vector (red arrow). As a result in this example the
physics-based reconstruction tilts the person at the torso to maintain stable pose given the
incorrect gravity vector (see the two bottom rows in \mbox{fig.~\ref{fig:pimfail} (c))}.

%% file: fig_h36m_qualitative.tex
\begin{figure}
  \begin{center}
    \includegraphics[width=0.8\linewidth]{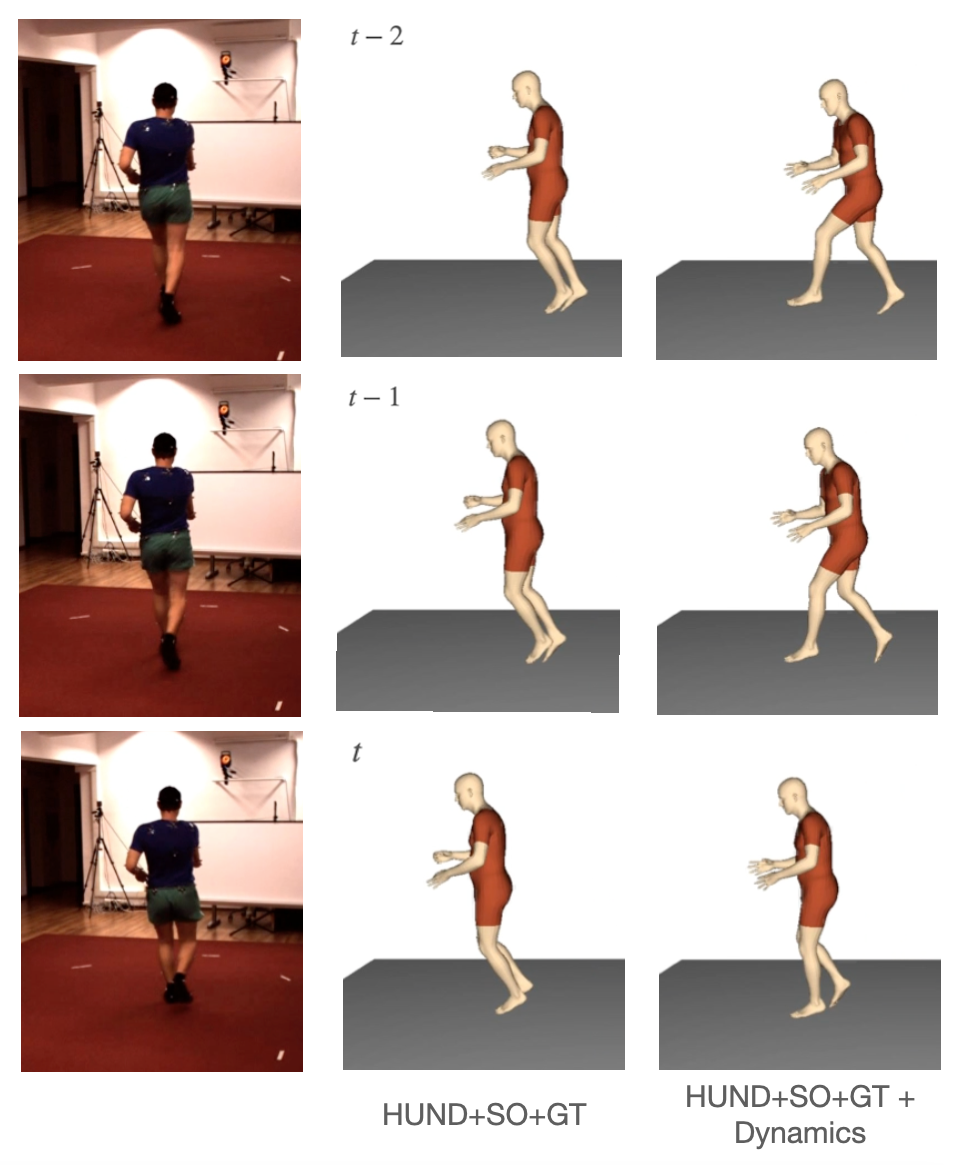}
    \vspace{-2mm}
  \caption{Qualitative results on the Human3.6M dataset. Note how the dynamical model (right) recovers plausible locomotion.}
    \label{fig:h36m_example}
  \end{center}
\vspace{-8mm}
\end{figure}

%% file: tab_ablation.tex
\begin{table*}[htp]
\begin{small}
\begin{center}
\scalebox{0.90}{
\begin{tabular}{l|c|c|c|c|c|c|c}
\textbf{Model} & \textbf{MPJPE-G} & \textbf{MPJPE} & \textbf{MPJPE-PA} & \textbf{MPJPE-2d} & \textbf{Velocity} & \textbf{Footskate (\%)} & \textbf{Float (\%)} \\
\hline
HUND+SO & 178 & 85 & 62 & 12 & 1.3 & 25 & 40 \\
HUND+SO + Dynamics & 167 & 87 & 62 & 12 & 0.45 & \textbf{7} & 1\\
HUND+SO+GT & 135 & 80 & 58 & 12 & 0.58 & 64 & 0\\
HUND+SO+GT + Dynamics & \textbf{132} & 80 & \textbf{57} & \textbf{11} & \textbf{0.27}  & 8 & 0  \\
\hline
HUND+SO+GT + Dynamics &  &  &  &  &  &  &  \\
\phantom{x} w/o 2d re-projection, \Eq{eq:keypointloss} & 154 &  104 & 68 & 17 & 0.32 & -  & - \\
\phantom{x} w/o 3d joints, \Eq{eq:quat_loss} & 134 & 84 & 60 & 11 & 0.27 & -  & -  \\
\phantom{x} w/o COM, \Eq{eq:com_loss} & 149 & 81 & 57 & 11 & 0.31 & -  &  - \\
\phantom{x} w/o COM and 3d joints, (\ref{eq:com_loss}, \ref{eq:quat_loss}) & 151 & 85 & 59 & 11 & 0.33 & -  & -  \\
\phantom{x} w/o pose prior, \Eq{eq:nfprior} & 138 & 80 & 57 & 11 & 0.24 & - & - \\
\end{tabular}
}
\end{center}
\vspace{-4mm}
\caption{Ablation experiments of the dynamics model on a validation set of 20 sequences from the Human3.6M dataset.}
\label{tab:dynablation}
\end{small}
\vspace{-1mm}
\end{table*}

%% file: tab_eval_humaneva_aist_h36m.tex
\begin{table*}[!htbp]
\begin{small}
\begin{center}
\scalebox{0.90}{
\begin{tabular}{c|l|c|c|c|c|c|c}
\textbf{Dataset} & \textbf{Model} & \textbf{MPJPE-G} & \textbf{MPJPE} & \textbf{MPJPE-PA} & \textbf{MPJPE-2d} & \textbf{Velocity} & \textbf{Footskate (\%)} \\
\hline
\multirow{8}{*}{Human3.6M} & VIBE \cite{kocabas20cvpr} & 208 & 69 & 44 & 16 & 0.32 & 27 \\
& PhysCap \cite{PhysCapTOG2020} & - & 97 & 65 & - & - & -  \\
& SimPoE~\cite{yuan2021simpoe} & - & \textbf{57} & \textbf{42} & - & -  \\
& Shimada \etal \cite{PhysAwareTOG2021}  & - & 77 & 58 & - & - & -  \\
& Xie \etal \cite{xie2021iccv} (Kinematics) & - & 74 & - & - & - & -  \\
& Xie \etal \cite{xie2021iccv} (Dynamics) & - & 68 & - & - & - & -  \\
& Ours: HUND+SO+GT & 145 &  83 & 56 & 14 & 0.46 & 48 \\
& Ours: HUND+SO+GT + Dynamics & \textbf{143} & 84 & 56 & \textbf{13} & \textbf{0.24} & \textbf{4}  \\
\hline
\multirow{4}{*}{HumanEva-I} & Rempe et al.~\cite{RempeContactDynamics2020} (Kinematics) & 408 &- &- & - & - & -\\
& Rempe et al.~\cite{RempeContactDynamics2020} (Dynamics) & 422 & -& - & - & - & - \\
& Ours: HUND+SO+GT  & 208 & \textbf{90} & 76 & 14 & 0.51 & 40 \\
& Ours: HUND+SO+GT + Dynamics & \textbf{196} & 91 & \textbf{74} & 14 & \textbf{0.27} & \textbf{4} \\

\hline
\multirow{2}{*}{AIST} 
& Ours: HUND+SO+GT & 156 & \textbf{107} & \textbf{67} & \textbf{10} & 0.59 & 51\\
& Ours: HUND+SO+GT + Dynamics  & \textbf{154} & 113 & 69 & 13 & \textbf{0.41} & \textbf{4}
\end{tabular}
}
\end{center}
\vspace{-4mm}
\caption{Quantitative results of our models compared to prior work on Human3.6M~\cite{h36m_pami}, \mbox{HumanEva-I}~\cite{Sigal:IJCV:10b} and a subset of AIST~\cite{li2021aistplusplus,aist-dance-db}.}
\label{t:quantitative-all}
\label{quantiative-asit}
\label{quantiative-h36m}
\label{t:quantitative-humaneva}
\vspace{-4mm}
\end{small}
\end{table*}

%% file: fig_pim_qualitative_arxiv.tex
\begin{figure*}
  \begin{center}
    \includegraphics[width=1\linewidth]{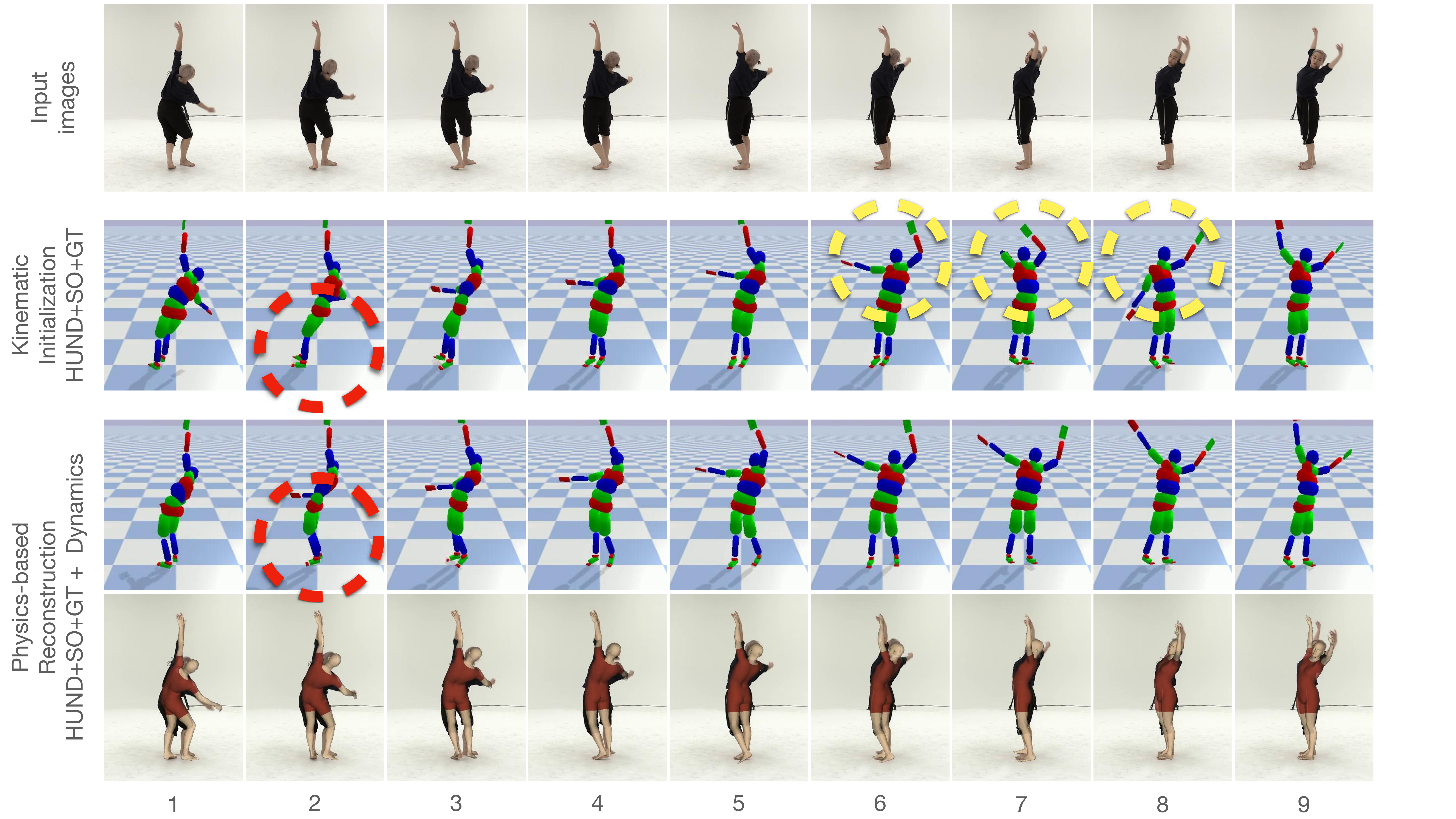}
  \vspace{-4mm}
  \caption{Example result on AIST~\cite{aist-dance-db}.
    The kinematic initialization produces poses that are unstable in the presence of gravity (red circle) or poses that are temporally inconsistent (yellow circles). Our physics-based approach corrects both errors.}
    \label{fig:aistqualitative}
    \vspace{-8mm}
  \end{center}
\end{figure*}

\begin{figure*}
  \begin{center}
    \begin{tabular}{l}
      \includegraphics[width=1.0\linewidth]{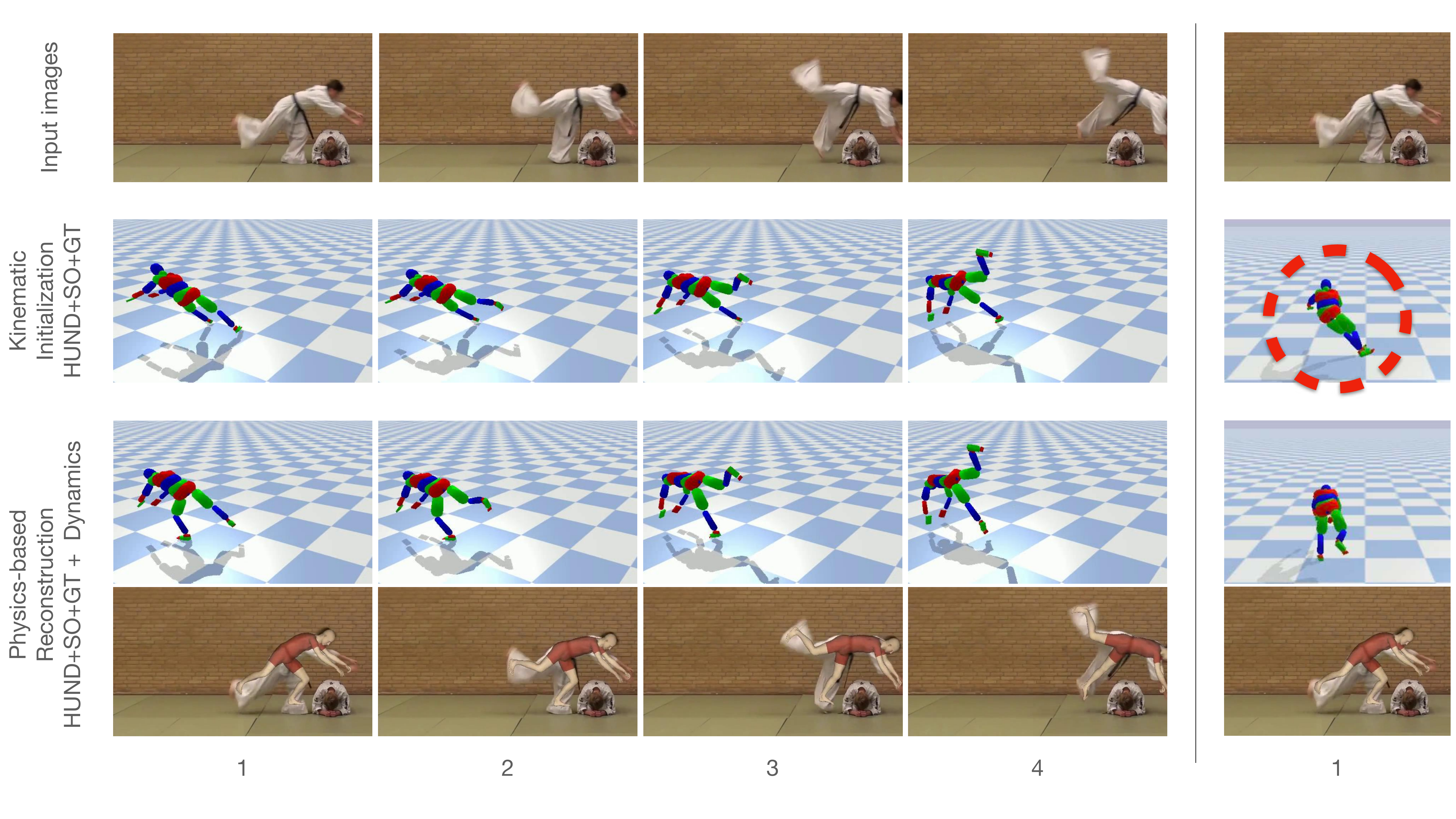} \\
      \includegraphics[width=1.0\linewidth]{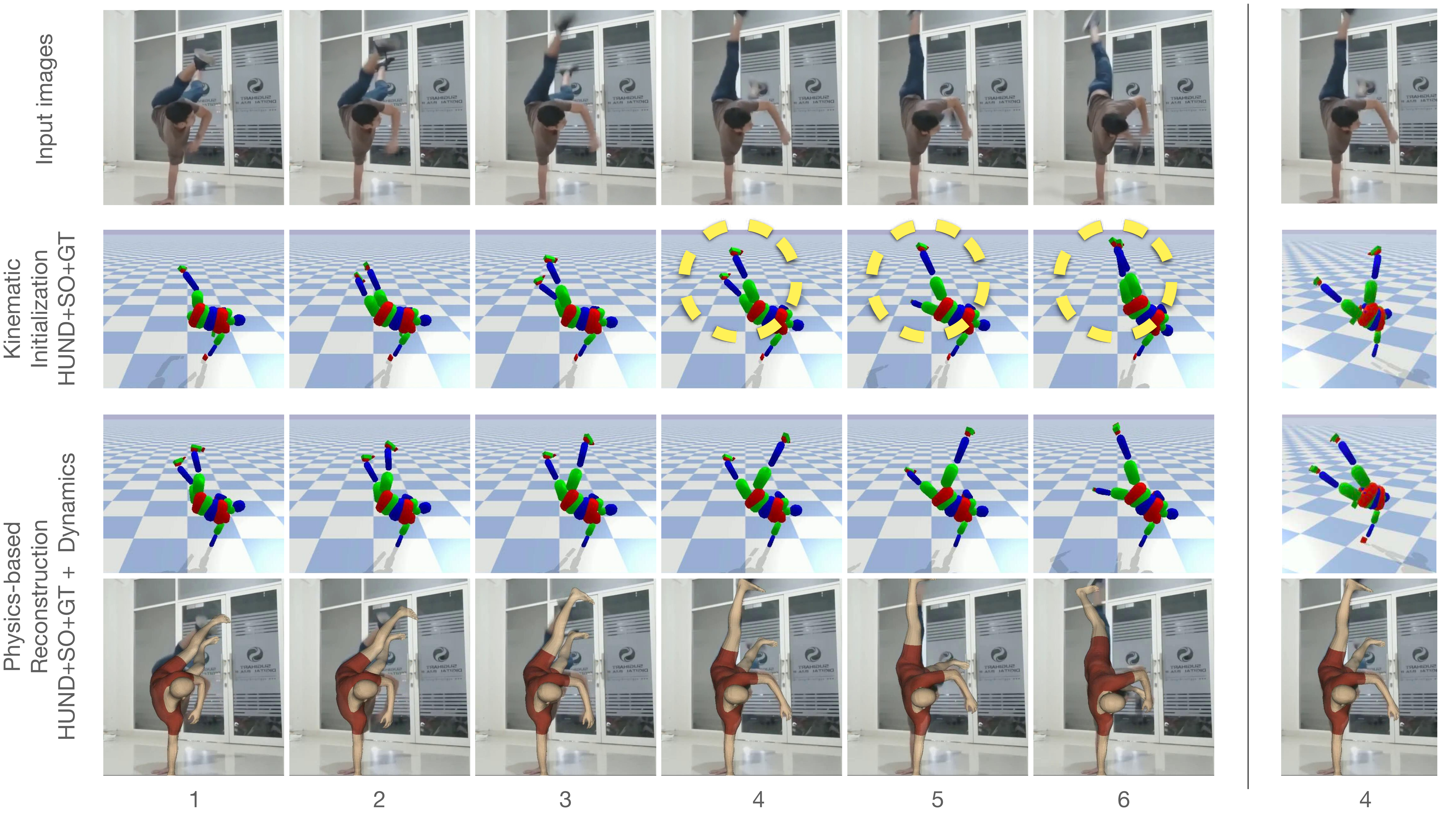}\\
      \end{tabular}
  \vspace{-4mm}
  \caption{Example results on real-world videos. In the top row sequence, the kinematic initialization incorrectly places the left foot before the jump. We highlight the mistake by showing the scene from another viewpoint (red circle). The kinematic initialization also fails to produce temporally consistent poses in the example in the bottom row (yellow circles). Our physics-based inference corrects both errors and generates a more plausible motion. See \href{https://tiny.cc/traj-opt}{tiny.cc/traj-opt}
    for more results.}
    \label{fig:pimqualitative}
  \end{center}
\end{figure*}

\begin{figure*}
  \begin{center}
    \includegraphics[width=1.12\linewidth]{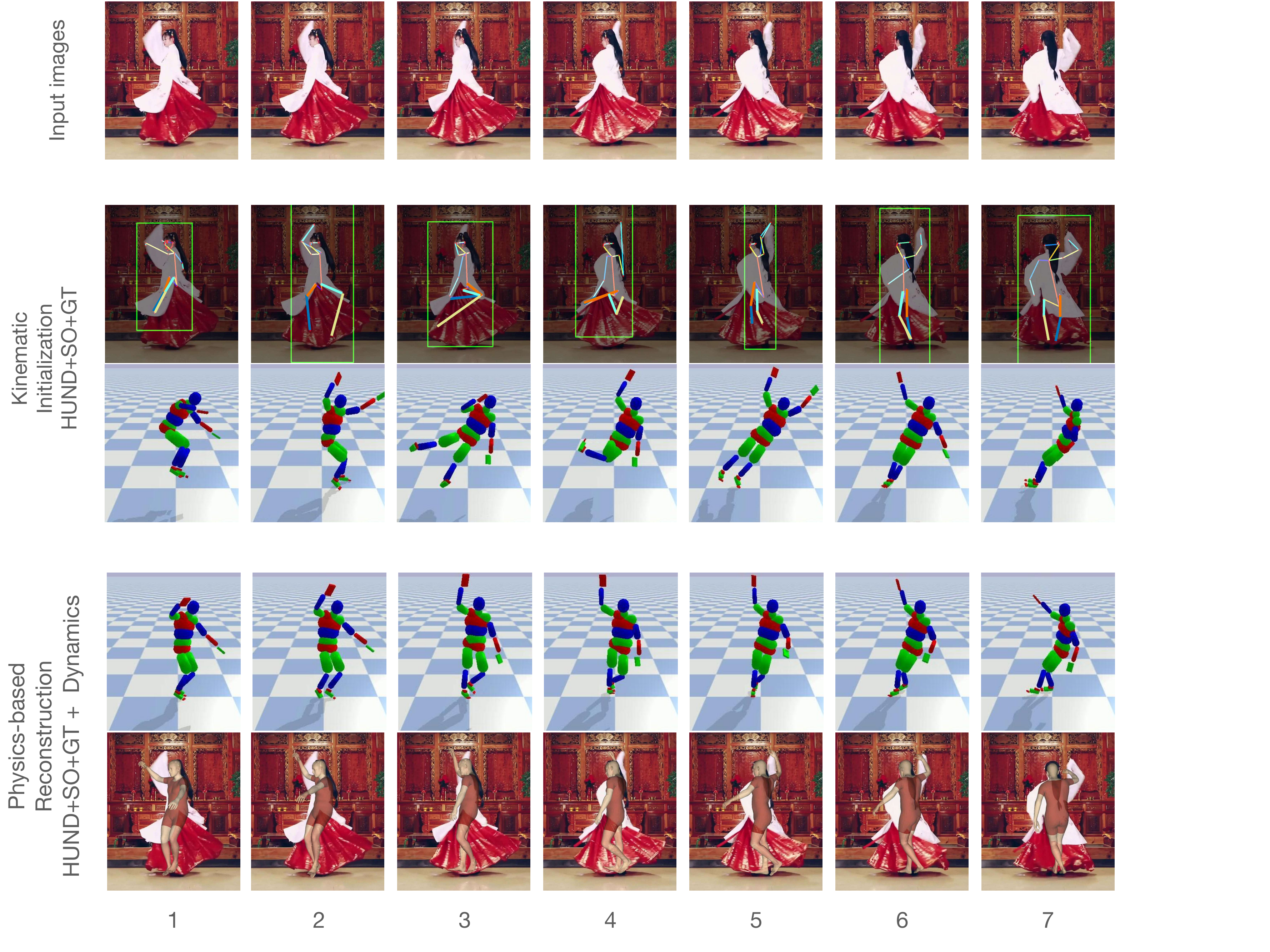}
  \caption{Example results on a difficult real-world video in which the legs of the person are occluded
    by the clothing. Note that 2d keypoints on the legs are incorrectly localized in multiple
    consecutive frames due to severe
    occlusion (second row) which results in poor 3d pose estimation by the kinematic model (third
    row). Interestingly our full model with dynamic is able to recover from errors in the kinematic
    initialization and generates reasonable sequence of 3d body poses (fourth row).}
    \label{fig:pimdress}
  \end{center}
\end{figure*}

\begin{figure}
  \begin{center}
    \includegraphics[width=1.01\linewidth]{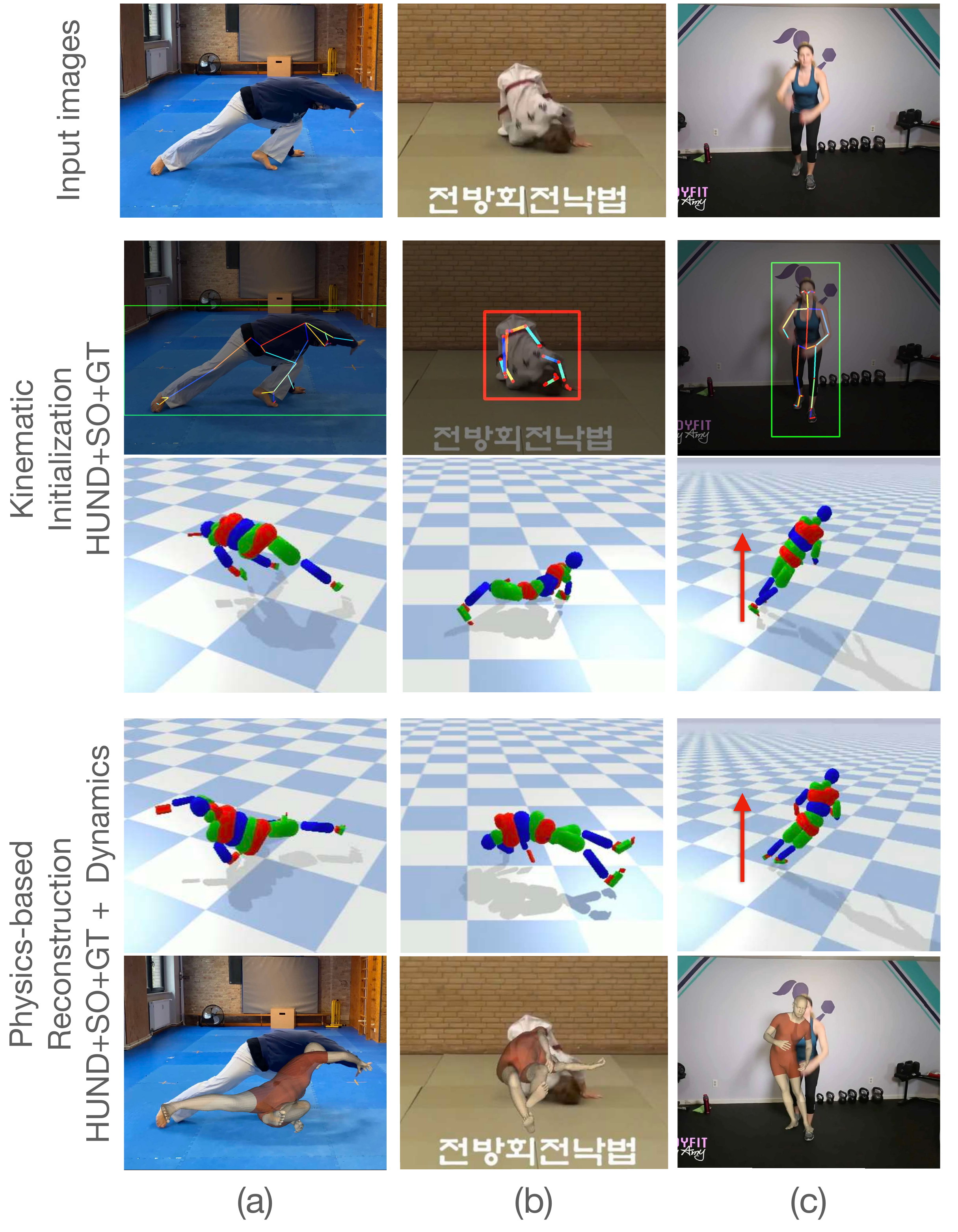}
    \caption{Examples of the characteristic failure cases of our approach on the real-world videos.
      Note that physics-based modeling introduces additional coupling between positions of the body
      limbs. While this is typically seen as an advantage, it also means that failure to estimate
      one limb correctly can propagate to other body limbs. For example in (a)
      our approach failed to correctly estimate position of the left arm which is used to support
      the body. As a result the overall 3d pose is worse for the dynamics (forth row) compared to
      the kinematic initialization (third row). Our physics-based reconstruction might also fail due to poor 
      kinematics initialization (b) or due to failure to correctly estimate the
      orientation of the ground plane relative to the camera (c).}
    \label{fig:pimfail}
  \end{center}
\end{figure}

%% file: conclusion.tex
\section{Conclusion}
In this paper, we have proposed a physics-based approach to 3d articulated video reconstruction of humans. By closely combining kinematic and dynamic constraints within an optimization process that is contact, mass, and inertia aware, with values informed by body shape estimates, we are able to improve the physical plausibility and reduce reconstruction artifacts compared to purely kinematic approaches. One of the primary goals of our work has been to demonstrate the advantages of incorporating an expressive physics model into the 3d pose estimation pipeline. Clearly, such a model makes inference more involved compared to specialized physics-based approaches such as \cite{PhysCapTOG2020,xie2021iccv}, but with the added benefit of being more capable and general.

\myparagraph{Ethical considerations.} This work aims to improve the quality of human pose reconstruction through the inclusion of physical constraints. We believe that the level of detail in our physical model limits its applications in tasks such as person identification or surveillance. The same limitation also prevents its use in the generation of e.g. deepfakes, particularly as the model lacks a photorealistic appearance. We believe our model is inclusive towards and supports a variety of different body shapes and sizes. While we do not study this in the paper, we consider it important future work. 

\myparagraph{Acknowledgements.} We would like to thank Erwin Coumans for his help with the project, as well as the supportive anonymous reviewers for their insightful comments.

%% file: supplementary.tex
\appendix
\section*{Appendix}

\noindent This supplementary material provides further details on our methodology and the data we used. \Section{sec:shape_approximation} presents details on our physical human body model, \Section{sec:sim_details} provides details regarding our simulation parameters, \Section{sec:metrics} presents our physics metrics, in \Section{sec:datasets} we present the datasets used in our experiments, \Section{sec:hyperparameters} provides details about our method's hyperparameters, and lastly \Section{sec:computation} summarizes our computational setup. When referring to equations or material in the main paper we will denote this by \mpp. Finally, please see our supplemental video for qualitative results of our method at \href{https://tiny.cc/traj-opt}{tiny.cc/traj-opt}.

\section{Physical Body Model}\label{sec:shape_approximation}
\begin{figure}[tbhp]
  \centering
  \includegraphics[width=1.0\linewidth]{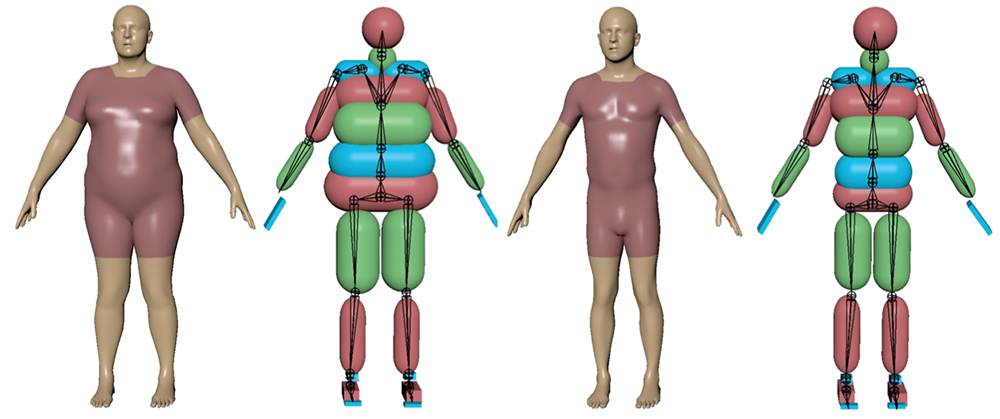}
  \caption{The physical body model's shape and mass parameters are based on an associated GHUM~\cite{xu2020ghum} mesh.}
  \label{fig:shape_approximation}
\end{figure}

Given a GHUM~\cite{xu2020ghum} body mesh $\MM(\bbeta, \btheta_0)$ associated with the shape parameters $\bbeta$ and the rest pose $\btheta_0$, we build a simulation-ready rigid multibody human model that best approximates the mesh with a set of parameterized geometric primitives (\cf \Figure{fig:shape_approximation}). 
The hands and feet are approximated with boxes whereas the rest of the body links are approximated with capsules. 
The primitives are connected and articulated with the GHUM body joints.

Inspired by \cite{alborno18cgf}, we optimize the primitive parameters by minimizing 
\begin{align}
    L(\bpsi) =& \sum_{b \in \BB} \sum_{\vv_g\in \MM_b} \min_{\vv_p \in \hat \MM_b}||\vv_g - \vv_p|| +\notag\\
    +& \sum_{b \in \BB} \sum_{\vv_p \in \hat\MM_b} \min_{\vv_g \in \MM_b}||\vv_p - \vv_g||,
\label{eq:shape_approx_loss}
\end{align}
where $\bpsi$ are the size parameters for the primitives, i.e. length and radius for the capsules, and depth, height and width for the boxes. The loss penalizes the bi-directional distances between pairs of nearest points on the GHUM mesh $\MM_b$ and surface of the primitive geometry $\hat \MM_b$ associated with the body link $b$.   

Furthermore, we learn a nonlinear regressor $\bpsi(\bbeta)$ with an MLP that performs fast shape approximation at run time.  
The regressor consists of two $256$-dimensional fully connected layers, and is trained with $50$K shapes generated with Gaussian sampling of the latent shape space $\bbeta$ together with the paired optimal primitive parameters using \Eq{eq:shape_approx_loss}.

Our physical model share an identical skeleton topology with GHUM but does not model the face and finger joints, due to the focused interest on the body dynamics in this work. Extending with finger joints, however, would enable simulation of hand-object interactions which would be interesting, but we leave this for future work. We note that there is a bijective mapping for the shared $16$ body joints between our model and GHUM, which allows for fast conversion between the physical and stastical representation.

\begin{table*}[htb]
    \centering
    \begin{tabular}{l|c|c|c|c}
        \textbf{Weight} & \textbf{H36M} & \textbf{AIST} & \textbf{HumanEva-I} & \textbf{Grid} \\
        \hline
         $w_{COM}$ & 15.0 & 15.0 &  15.0 & \{1, 2, 5, 10, 15, 25 \} \\
         $w_{pose}$ & 0.5 & 0.5 & 0.5 & \{0.1, 0.5, 1, 2 \} \\ 
         $w_{2d}$ & 4.0 & 4.0 & 4.0 & \{1, 2, 4, 8, 10 \} \\
         $w_{nf}$ & 1.0 & 1.0 & 1.0 & \{0.001, 0.1, 1, 10\} \\ 
         $w_{TV}$ & 1.0 & 1.0 & 1.0 & \{0.1, 1, 10\} \\ 
         $w_{lim}$ & 1.0 & 1.0 & 1.0 & \{0.1, 1, 10\} \\
    \end{tabular}
    \caption{Weights of the objective function described in \S3.3~\mpp and \Eq{eq:weighted_loss} for our three main datasets: Human3.6M~\cite{h36m_pami}, AIST~\cite{aist-dance-db}, and HumanEva-I~\cite{Sigal:IJCV:10b}. ``Grid'' specifies the values evaluated while selecting hyperparameter values. Note that we did not exhaustively explore all combination.}
    \label{tab:hyperparams}
\end{table*}

\section{Simulation Details}\label{sec:sim_details}
We run the Bullet simulation at $200$ Hz, with friction coefficient $\mu=0.9$  and gravitational acceleration constant $9.8~\textrm{m/s}^2$. The PD-controllers controlling each torque motor is tuned with position gain $k_p = 4.0$, velocity gain $k_d = 0.3$, and torque limits similar to those presented in \cite{2018-TOG-deepMimic}.

\begin{table}[tb]
    \centering
    \begin{tabular}{c|c|c|c}
        \textbf{Sequence} & \textbf{Subject} &
        \textbf{Camera Id} & \textbf{Frames}  \\
        \hline
        Phoning & S11 & 55011271 & 400-599 \\
        Posing\_1 & S11 & 58860488 & 400-599 \\
        Purchases & S11 & 60457274 & 400-599 \\
        SittingDown\_1 & S11 & 54138969 & 400-599 \\
        Smoking\_1 & S11 & 54138969 & 400-599 \\
        TakingPhoto\_1 & S11 & 54138969 & 400-599 \\
        Waiting\_1 & S11 & 58860488 & 400-599 \\
        WalkDog & S11 & 58860488 & 400-599 \\
        WalkTogether & S11 & 55011271 & 400-599 \\
        Walking\_1 & S11 & 55011271 & 400-599 \\
        Greeting\_1 & S9 & 54138969 & 400-599 \\
        Phoning\_1 & S9 & 54138969 & 400-599 \\
        Purchases & S9 & 60457274 & 400-599 \\
        SittingDown & S9 & 55011271 & 400-599 \\
        Smoking & S9 & 60457274 & 400-599 \\
        TakingPhoto & S9 & 60457274 & 400-599 \\
        Waiting & S9 & 60457274 & 400-599 \\
        WalkDog\_1 & S9 & 54138969 & 400-599 \\
        WalkTogether\_1 & S9 & 55011271 & 400-599 \\
        Walking & S9 & 58860488 & 400-599
    \end{tabular}
    \vspace{2mm}
    \caption{The subset of Human3.6M used in the ablation experiments. Note that the data was downsampled from 50 to 25 FPS.}
    \label{tab:validation_h36m_sequences}
\end{table}

\begin{table}[tb]
     \centering
     \begin{tabular}{c|c}
         \textbf{Sequence} & \textbf{Frames}  \\
         \hline
gBR\_sBM\_c06\_d06\_mBR4\_ch06 & 1-120 \\
gBR\_sBM\_c07\_d06\_mBR4\_ch02 & 1-120 \\
gBR\_sBM\_c08\_d05\_mBR1\_ch01 & 1-120 \\
gBR\_sFM\_c03\_d04\_mBR0\_ch01 & 1-120 \\
gJB\_sBM\_c02\_d09\_mJB3\_ch10 & 1-120 \\
gKR\_sBM\_c09\_d30\_mKR5\_ch05 & 1-120 \\
gLH\_sBM\_c04\_d18\_mLH5\_ch07 & 1-120 \\
gLH\_sBM\_c07\_d18\_mLH4\_ch03 & 1-120 \\
gLH\_sBM\_c09\_d17\_mLH1\_ch02 & 1-120 \\
gLH\_sFM\_c03\_d18\_mLH0\_ch15 & 1-120 \\
gLO\_sBM\_c05\_d14\_mLO4\_ch07 & 1-120 \\
gLO\_sBM\_c07\_d15\_mLO4\_ch09 & 1-120 \\
gLO\_sFM\_c02\_d15\_mLO4\_ch21 & 1-120 \\
gMH\_sBM\_c01\_d24\_mMH3\_ch02 & 1-120 \\
gMH\_sBM\_c05\_d24\_mMH4\_ch07 & 1-120
     \end{tabular}
     \vspace{2mm}
     \caption{Sequences used for evaluation on AIST.}
 \label{tab:aist_sequences}
\end{table}

\begin{table}[ht]
    \centering
    \begin{tabular}{c|c|c}
        \textbf{Sequence} & \textbf{Subject} &
        \textbf{Camera Id} \\
        \hline
        S11 & Directions\_1 & 60457274 \\
        S11 & Discussion\_1 & 60457274 \\
        S11 & Greeting\_1 & 60457274 \\
        S11 & Posing\_1 & 60457274 \\
        S11 & Purchases\_1 & 60457274 \\
        S11 & TakingPhoto\_1 & 60457274 \\
        S11 & Waiting\_1 & 60457274 \\
        S11 & WalkDog\_1 & 60457274 \\
        S11 & WalkTogether\_1 & 60457274 \\
        S11 & Walking\_1 & 60457274 \\
        S9 & Directions\_1 & 60457274 \\
        S9 & Discussion\_1 & 60457274 \\
        S9 & Greeting\_1 & 60457274 \\
        S9 & Posing\_1 & 60457274 \\
        S9 & Purchases\_1 & 60457274 \\
        S9 & TakingPhoto\_1 & 60457274 \\
        S9 & Waiting\_1 & 60457274 \\
        S9 & WalkDog\_1 & 60457274 \\
        S9 & WalkTogether\_1 & 60457274 \\
        S9 & Walking\_1 & 60457274
    \end{tabular}
    \vspace{2mm}
    \caption{The evaluation subset of Human3.6M used in the main evaluation. The subset is similar to the one used in \cite{PhysCapTOG2020}. We downsampled the data from 50 FPS to 25 FPS.}
    \label{tab:full_h36m_sequences}
\end{table}

\section{Additional Metrics}\label{sec:metrics}
In addition to the standard 2d and 3d joint position error metrics, we evaluate our reconstructions using physical plausibility metrics similar to those proposed in~\cite{RempeContactDynamics2020}. Since the authors were unable to share their code we implement our own versions the metrics which doesn't require foot-ground contact annotations. A foot contact is defined as at least $N=10$ vertices of a foot mesh being in contact with the ground plane. We set the contact threshold to $d=0.005$~m for kinematics. To account for the modeling error when approximating the foot with a box primitive we set the contact threshold for dynamics to $d=-0.015$~m. \\
\myparagraph{Footskate.} The percentage of frames in a sequence where either foot joint moves more than $2$ cm between two adjacent frames while the corresponding foot was in contact with the ground-plane. \\
\myparagraph{Float.} The percentage of frames in a sequence where at least one of the feet was not in contact but was within $2$ cm of the ground-plane. This metric captures the common issue of reconstructions floating above the ground while not penalizing correctly reconstructed motion of e.g. jumps. \\
\myparagraph{Velocity.} The mean error between the 3d joint velocities in the ground-truth data and the joint velocity in the reconstruction. High error velocity indicates that the estimated motion doesn't smoothly follow the trajectory of the true motion.
We define the velocity error as
\begin{equation}\label{eq:velocity_error}
    e_{v} = \frac{1}{N}\sum_{i=1}^N\sum_{k\in K} |\dot{\bar{\xx}}_k^i - \dot{{\xx}}_k^i|,
\end{equation}
where $\dot{\bar{\xx}}_k^i$ is the magnitude of the ground-truth 3d joint velocity vector (in m/s) for joint $k$ at frame $i$ and where $\dot{{\xx}}_k^i$ denotes the reconstructed joint. We estimate the velocity using finite differences from 3d joint positions and use first frame translation aligned joint estimates (as in MPJPE-G).

\section{Datasets}\label{sec:datasets}
\myparagraph{Human3.6M.} We use two subsets for our experiments on Human3.6M~\cite{h36m_pami}. When we compare our method to state-of-the-art methods we use a dataset split similar to the one used in \cite{xie2021iccv}. See \Table{tab:full_h36m_sequences} for the complete lists of sequences we use. Similarly to \cite{PhysCapTOG2020,xie2021iccv}, we down sample the sequences from 50 FPS to 25 FPS.

When perform ablations of our model we a smaller subset where we select 20 $4$-sec sequences from the test split of Human3.6M dataset (subjects 9 and 11). We selected sequences that show various dynamic motions such as walking dog,  running and phoning (with large motion range), to sitting and purchasing (with occluded body parts). For each sequence, we randomly selected one of the four cameras. We list the sequences in~\Table{tab:validation_h36m_sequences}. 

\myparagraph{HumanEva-I.} We evaluate our method on the subset of  HumanEva-I walking sequences~\cite{Sigal:IJCV:10b} as selected by~\cite{RempeContactDynamics2020}, see \Table{tab:he_sequences}.

\myparagraph{AIST.} We select four second video sequences from the public dataset~\cite{aist-dance-db,li2021aistplusplus}, showing fast and complex dancing motions, picked randomly from one of the $10$ cameras. We list our selected sequences in \Table{tab:aist_sequences}.

\myparagraph{''In-the-wild" internet videos.} We perform qualitative evaluation of our model on videos of dynamic motions rarely found in laboratory captured datasets. These videos were made available on the internet under a CC-BY license which grants the express permission to be used for any purpose. Note that we only used the videos to perform qualitative analysis of our approach -- the videos will not be redistributed as a dataset.

\begin{table}[ht]
     \centering
     \begin{tabular}{c|c|c|c}
         \textbf{Sequence} & \textbf{Subject} &
     \textbf{Camera Id} & \textbf{Frames}  \\
         \hline
         Walking & S1 & C1 & 1-561 \\
         Walking & S2 & C1 & 1-438 \\
         Walking & S3 & C1 & 1-490
     \end{tabular}
     \vspace{2mm}
     \caption{Sequences used for evaluation on HumanEva-I.}
 \label{tab:he_sequences}
\end{table}

\subsection{Human Data Usage}
This work relies on recorded videos of humans. Our main evaluation is performed on two standard human pose benchmarks: Human3.6M\footnote{\url{http://vision.imar.ro/human3.6m/}}~\cite{h36m_pami} and AIST\footnote{\url{https://aistdancedb.ongaaccel.jp/}}~\cite{aist-dance-db}. These datasets have been approved for research purposes according to their respective websites. Both datasets contain recordings of actors in laboratory settings. To complement this, we perform qualitative evaluation on videos released on the internet under creative commons licenses. %

\section{Hyperparameters}\label{sec:hyperparameters}
The most important hyperparameters are the weights of the weighted objected function described in \S3.3~\mpp. Where combined loss function is given by

\begin{align}\label{eq:weighted_loss}
    \begin{split}
    L &= w_{COM} L_{COM} + w_{pose} L_{pose}  \\ 
        &+ w_{2d} L_{2d}  + w_{nf} L_{nf} + w_{TV} L_{TV}  \\ 
        &+ w_{lim} L_{lim}.
\end{split}
\end{align}

We tuned the weights on sequences from the training splits. The goal was to scale the different components such that they have roughly equal magnitudes while minimizing the MPJPE-G error. See \Table{tab:hyperparams} for details regarding the search grid and the chosen parameter values.

\section{Computational Resources}\label{sec:computation}
For running small experiments we used a desktop workstation equipped with an ``Intel(R) Xeon(R) CPU E5-2620 v4 @ 2.10GHz'' CPU, 128 GB system memory and two NVIDIA Titan Xp GPUs. We ran kinematics in the cloud using instances with a V100 GPU, 48 GB of memory and 8 vCPUs. In the dynamics experiments, we used instances with 100 vCPUs and 256 GB of memory for the CMA-ES~\cite{Hansen2006} optimization. Optimizing a window of 1 second of video takes roughly 20 min using a 100 vCPUs instance.